\documentclass[conference]{IEEEtran}
\IEEEoverridecommandlockouts
\usepackage{cite}
\usepackage{amsmath,amssymb,amsfonts}
\usepackage{algorithmic}
\usepackage{graphicx}
\usepackage{url}
\usepackage{tikz}
\usetikzlibrary{arrows.meta,backgrounds,fit,positioning}
\usepackage{booktabs}
\usepackage{textcomp}
\usepackage{xcolor}
\def\BibTeX{{\rm B\kern-.05em{\sc i\kern-.025em b}\kern-.08em
    T\kern-.1667em\lower.7ex\hbox{E}\kern-.125emX}}
\renewcommand{\footnoterule}{%
  \kern -3pt%
  \hrule width 0.33\columnwidth height 0.4pt%
  \kern 2.6pt%
}
\begin{document}

\title{GeoSEAN: Explainable Country-Level Image Geolocation for ASEAN Regions}

\author{\IEEEauthorblockN{1\textsuperscript{st} Muhamad Syukron}
\IEEEauthorblockA{\textit{Department of Statistics} \\
\textit{Institut Teknologi Sepuluh Nopember}\\
Surabaya, Indonesia \\
muhamad.syukron@its.ac.id}
\and
\IEEEauthorblockN{2\textsuperscript{nd} Danish Rafie Ekaputra}
\IEEEauthorblockA{\textit{Department of Statistics} \\
\textit{Institut Teknologi Sepuluh Nopember}\\
Surabaya, Indonesia \\
5052231024@student.its.ac.id}
\and
\IEEEauthorblockN{3\textsuperscript{rd} Tintrim Dwi Ary Widhianingsih}
\IEEEauthorblockA{\textit{Department of Statistics} \\
\textit{Institut Teknologi Sepuluh Nopember}\\
Surabaya, Indonesia \\
dwi.ary@its.ac.id}
}

\maketitle

\begin{abstract}
Image geolocation aims to infer the geographic origin of an image from visual content alone. However, this task remains challenging in regions where countries share similar urban, roadside, architectural, and environmental characteristics. Many existing geolocation models focus on coordinate level prediction or classification performance while providing limited insight into how visual evidence contributes to location predictions. This study presents an explainable country level image geolocation pipeline for 11 ASEAN countries. First, we collected 4,850 images from GeoGuessr style sources, Google Images, and additional street level imagery. We then evaluated three approaches on this dataset: CLIP zero shot classification, a LightGBM classifier, and an MLP classifier. The MLP achieved the best test performance, attaining an accuracy and F1 score of 85.91\%. For explainability, predictions generated by the MLP classifier were analyzed post hoc using CLIP attention rollout, YOLO26 object detection on the original images, and Energy Based Pointing Game (EBPG) overlap metrics. Object level analysis indicates that frequently detected objects are not necessarily associated with the highest attention density, suggesting that object frequency and attention based visual evidence capture different aspects of a scene. These results demonstrate that the proposed model can support accurate regional image geolocation while enabling object level inspection of the visual cues underlying its predictions.
\end{abstract}


\section{Introduction}

Understanding the geographic origin of an image is an important capability for both humans and intelligent systems. Given a single image, people often want to know where it was captured, such as the country or region it represents. However, this task is inherently challenging because many regions share highly similar visual appearances \cite{fang2026geomr, chen2025enhancing}. For example, cities in Southeast Asia often exhibit comparable architectural styles, road structures, and urban environments. These visual similarities make it difficult even for humans to reliably infer geographic origin from appearance alone.

This challenge becomes increasingly critical in real world digital environments, where large volumes of images are shared without reliable metadata, leaving users without explicit location information. On many online platforms, such as image sharing services like Instagram and Pinterest, images are frequently uploaded without geographic tags. In many cases, users may wish to visit these locations for travel but are unable to determine where an image was captured. This limitation highlights the growing need for image based geolocalization methods that can infer geographic information directly from visual content.

Existing research in image based geolocalization has achieved significant progress in improving global scale prediction accuracy (e.g. \cite{He, Zhu, Weyand, Clark, Haas}). Most approaches focus on learning discriminative visual representations to estimate geographic coordinates across the world. However, these methods are often designed as black box systems, making it difficult for users to understand or verify why a particular location is predicted. Without interpretability, the reliability and usability of these systems in real world applications remain limited.

A recent work by Theiner et al. \cite{Theiner} attempts to address interpretability in geolocalization tasks. They employ a CNN \cite{lecun1998gradient} for classification, where each image is assigned to a geographic cell. They further compute Concept Influence (CI), defined as the gradient of the CNN combined with a segmentation model, to identify the objects that most influence the predicted location. However, this approach relies on conventional architectures and may lead to suboptimal performance.

To address this limitation, we use a frozen CLIP vision transformer as an image encoder and train an MLP classifier for country-level geolocation. The attention scores produced by the transformer are then leveraged to construct attention maps, which are later used to analyze object-level influence. In addition, object detection is performed independently on the original image, and the detected bounding boxes are compared with the attention maps to identify objects that receive high attention during prediction.

To the best of our knowledge, this is the first work that combines image classification and object detection tasks to infer the geographic location of an image while also providing an explanation of what influences the prediction. Through this approach, users can not only determine where an image was taken but also understand why the model produces a particular location prediction.

Furthermore, publicly available geolocation datasets for ASEAN countries remain limited, making image geolocation particularly challenging in this region. To address this gap, we introduce the ASEAN Geolocation dataset, a new benchmark designed specifically for country level geolocation across ASEAN nations. We anticipate that this dataset will facilitate future research and contribute to the development of more accurate geolocation models for the region.

\section{Related Work}

He et al. \cite{He} propose a geolocation method using CNNs to match ground-level and aerial images. Zhu et al. \cite{Zhu} extend this idea by replacing CNNs with a Transformer-based architecture \cite{vaswani2017attention}. However, these approaches require both viewpoints, which limits its applicability during inference when only ground-level images are available.

Several studies focus on ground-level images only. Weyand et al. \cite{Weyand} formulate geolocation as a classification problem by dividing the world into 26,263 spatial cells and training a CNN to predict the corresponding cell. Pramanick et al. \cite{Pramanick} follow a similar formulation but adopt a Transformer architecture instead of CNNs. However, this fine-grained discretization introduces scalability challenges, increases computational cost, and leads to sparse class distributions.

To address the limitations of direct cell classification, later approaches introduce hierarchical and multi-stage frameworks. Clark et al. \cite{Clark} first classify street-view images into 16 semantic scene categories (e.g., forest, cultural, industrial) and then predict location across seven hierarchical geographic levels using a Swin Transformer \cite{liu2021swin}. Similarly, Haas et al. \cite{Haas} perform hierarchical geolocation, but rely on a more complex pipeline combining CLIP-based \cite{radford2021clip} representations with OPTICS clustering \cite{ankerst1999optics}.

Beyond classification based methods, regression based approaches aim to directly predict geographic coordinates. Cepeda et al. \cite{Cepeda} introduce GeoCLIP, which encodes images using CLIP and maps them to continuous longitude and latitude representations through a location encoder. In contrast to cell based classification methods, this approach enables direct coordinate regression. Building on this paradigm, RANGE \cite{Dhakal} was proposed as a retrieval based framework that synthesizes location representations from multiple visually similar regions instead of relying on direct contrastive alignment. Other studies explore alternative architectures for coordinate prediction, including EfficientNet \cite{KordopatisZilos2021}, image retrieval combined with large language models \cite{Jia_G3, Zhou}, and vision language models \cite{Jia_GeoRanker}.  

Predicting precise longitude and latitude at a global scale remains highly challenging. Therefore, several studies constrain the task to a coarser granularity, such as city level prediction, to improve reliability. For instance, Haas et al. \cite{HaasAlbertiSkreta2023} perform city level classification using CLIP in a zero shot setting, while \cite{WuHuang2022} extend this approach by incorporating few shot learning and fully supervised linear probing.  

Despite these advances, most existing methods primarily focus on improving geolocation accuracy, while interpretability remains underexplored. Current models often operate as black boxes and provide limited insight into the rationale behind their predictions. To address this limitation, Theiner et al. \cite{Theiner} propose an explainability framework that identifies influential image regions through segmentation and predicts contributing semantic classes. However, their approach relies on convolutional architectures and does not fully exploit attention mechanisms. In this work, we use a frozen CLIP vision transformer \cite{radford2021clip} for country-level classification and apply YOLO26 \cite{yolo26_ultralytics} independently to the original image for object detection. The detected objects are then analyzed against transformer attention maps to provide object-level explanations for the predicted geographic location.

\section{Methodology}

\subsection{Dataset}

\textbf{Dataset Collecion. }Publicly available geolocation datasets for ASEAN countries are limited. The most relevant dataset, a GeoGuessr style dataset available on Kaggle~\cite{kaggle_geoguessr_countries}, contains fewer than 500 images from ASEAN countries. Coverage is particularly sparse for several nations, with only two images from Myanmar and no images from Brunei or Timor Leste (see Table~\ref{tab:dataset_distribution} for the complete distribution). This scarcity of data presents a significant challenge for developing and evaluating geolocation models in the region.

To address this limitation, we introduce the ASEAN Geolocation dataset, which contains approximately 500 images for each ASEAN country to ensure sufficient coverage and reduce class imbalance. The dataset was collected from multiple publicly accessible sources, including GeoGuessr style images, Google Images, and street level imagery curated from Google Maps. The final dataset consists of 1,952 GeoGuessr style images, 1,573 Google Images, and 1,325 street level images. Table~\ref{tab:dataset_distribution} summarizes the country level distribution and the contribution of each data source. 

\begin{table}[!htbp]
\caption{ASEAN Geolocation dataset distribution by source.}
\label{tab:dataset_distribution}
\centering
\footnotesize
\setlength{\tabcolsep}{2.5pt}
\begin{tabular*}{0.82\columnwidth}{@{\extracolsep{\fill}}lrrrr}
\toprule
Country & GeoGuessr & Google & Street-level & Total \\
\midrule
Thailand & 400 & 100 & 0 & 500 \\
Singapore & 449 & 51 & 0 & 500 \\
Malaysia & 400 & 100 & 0 & 500 \\
Indonesia & 288 & 212 & 0 & 500 \\
Philippines & 219 & 281 & 0 & 500 \\
Cambodia & 118 & 232 & 0 & 350 \\
Myanmar & 2 & 134 & 214 & 350 \\
Laos & 61 & 45 & 394 & 500 \\
Vietnam & 15 & 225 & 260 & 500 \\
Brunei & 0 & 95 & 205 & 300 \\
Timor-Leste & 0 & 98 & 252 & 350 \\
\midrule
Total & 1,952 & 1,573 & 1,325 & 4,850 \\
\bottomrule
\end{tabular*}
\end{table}

\textbf{Dataset Labelling. }Each image is assigned a country-level label represented by an ISO-style country code, a country name, and an integer class label. GeoGuessr-style images inherit country labels from the original country-labeled collection. Google Images were collected using country-specific search queries that combine country names, major cities, street-scene terms, and locally distinctive visual cues, and their initial labels follow the target country of the query. To reduce labeling errors caused by irrelevant keyword matches, the authors manually verified the scraped Google Images for visual relevance, country consistency, and source context before inclusion. Additional street-level images were labeled using their location and country context, then filtered to remove corrupted, duplicate, irrelevant, or visually inconsistent samples. In the public version of the dataset, images are organized by country and filenames are made source-free to reduce the possibility that models learn shortcuts from file names or source identifiers rather than visual geographic cues.

\textbf{Ethics of Data Collection.} All images were collected from publicly accessible sources and were used only for research on country-level visual geolocation. Public web images and additional street-level images were used only for non-commercial research dataset construction, with filtering to remove irrelevant or unsuitable samples. The dataset does not use private user metadata for prediction, and the task label is limited to country of origin rather than personal identity or precise private information. During curation, images were filtered to remove unusable, duplicate, and irrelevant samples, and redistribution is limited to a research dataset intended to support reproducible evaluation by other researchers.

\textbf{Reproducibility.} To support future research in the ASEAN region, the curated dataset is publicly available through Hugging Face.\footnote{\url{https://huggingface.co/datasets/0xRafie/asean-geolocation}} The dataset contains labeled images organized by country, facilitating reproducible training and evaluation of image geolocation models.

\subsection{Experimental Design}
\begin{figure*}[t]
\centering
\includegraphics[width=0.92\textwidth]{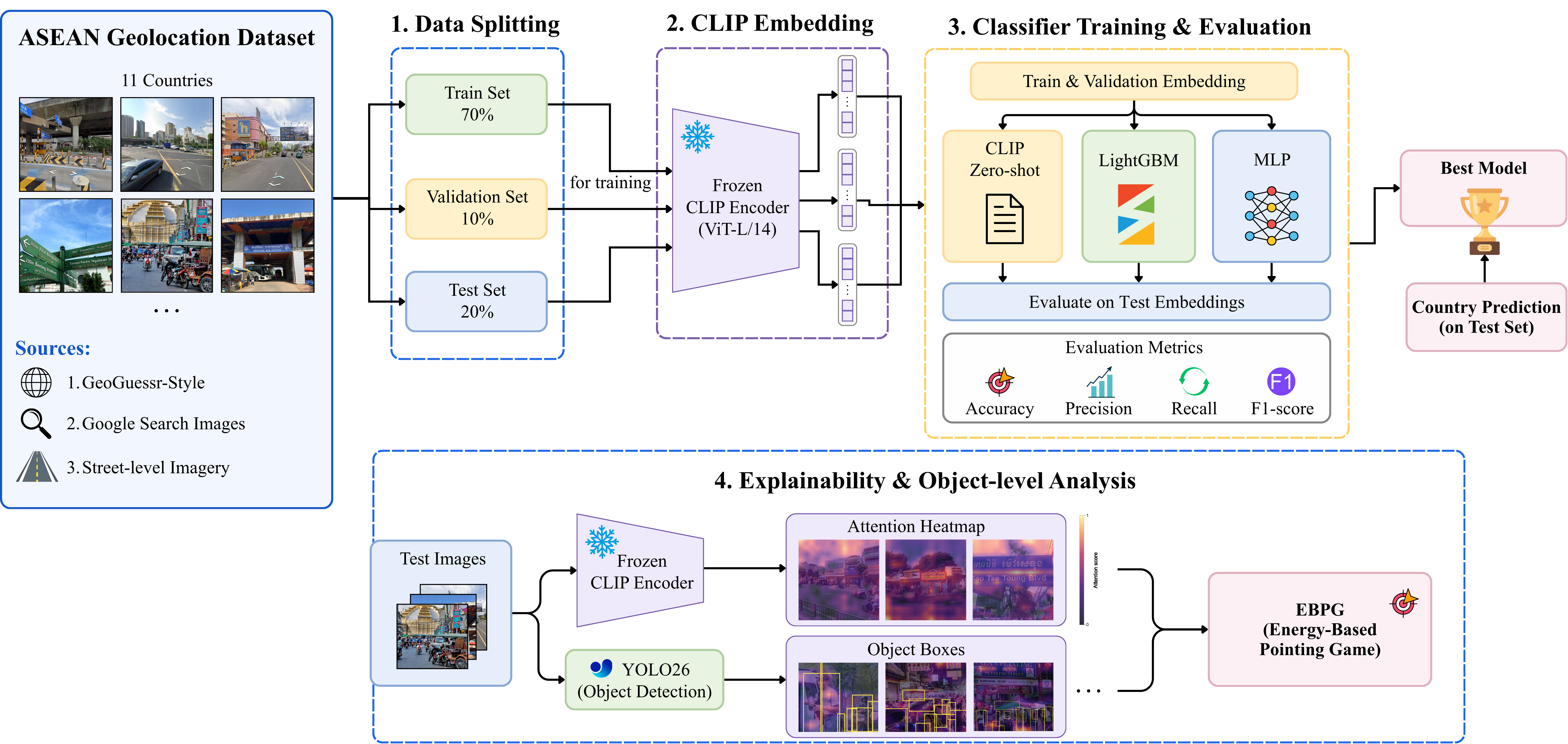}
\caption{Experimental workflow. The dataset is split before modeling. Training and validation subsets are used for classifier selection, while the held-out test subset is reserved for final evaluation. The post-hoc explanation branch uses only test images, where CLIP attention rollout and YOLO26 detections are compared using EBPG-based overlap measures.}
\label{fig:experimental_design}
\end{figure*}

This study begins with the ASEAN Geolocation dataset, which is divided into 70\% training, 10\% validation, and 20\% testing. Before feature extraction, each image is converted to RGB, resized and center-cropped to the CLIP ViT-L/14 input size of $224\times224$ pixels, rescaled to pixel values in $[0,1]$, and normalized using CLIP's pretrained channel mean and standard deviation. The preprocessed images are passed through a frozen CLIP vision encoder to extract visual embeddings. The frozen CLIP embeddings are used for classifier selection based primarily on validation accuracy, with validation F1-score used as a secondary criterion, while the held-out test set is used only for final evaluation.

For interpretability, the selected prediction pipeline is analyzed on test images. Transformer attention maps are generated from the CLIP vision encoder, while YOLO26 is applied independently to the original images to detect object bounding boxes. The attention maps and bounding boxes are then compared using EBPG-based overlap measures to support post-hoc object-level inspection and confidence-based analysis of incorrect predictions. The full experimental workflow is shown in Fig.~\ref{fig:experimental_design}.

\subsection{Image Embedding}
For image representation, this study uses the CLIP ViT-L/14 vision encoder \cite{radford2021clip} as a frozen feature extractor. CLIP was pretrained on large-scale image-text pairs, and its image encoder provides visual features that can be reused for downstream classification. In this study, the CLIP parameters are not updated. Keeping the encoder fixed isolates the contribution of the pretrained representation and avoids fitting a large vision backbone on a relatively small country-level dataset.

The encoder maps each preprocessed image to a global visual embedding, which is then used as the input representation for the country classifier. For an input image $\mathbf{x}_i$, this process is written as
\begin{equation}
\mathbf{z}_i = f_{\mathrm{CLIP}}(\mathbf{x}_i), \quad \theta_{\mathrm{CLIP}} \ \mathrm{fixed},
\end{equation}
where $f_{\mathrm{CLIP}}$ is the CLIP image encoder, $\theta_{\mathrm{CLIP}}$ denotes its fixed parameters, and $\mathbf{z}_i$ is the extracted image embedding.

\subsection{Classifier}
The classifier stage evaluates both zero-shot and supervised use of the frozen CLIP representation. For the zero-shot baseline, no dataset-specific classifier parameters are trained. Each ASEAN country is represented by five text prompts: ``a street-level photo in \{country\}.''; ``a street scene in \{country\}''; ``a geolocation photo from \{country\}''; ``a road scene in \{country\}''; and ``a photo taken in \{country\}.'' The prompt embeddings are normalized, averaged, and normalized again to form one text prototype per country. Each test image is encoded by CLIP, and the predicted class is the country prototype with the highest cosine similarity to the image embedding.

For supervised classification, the CLIP image encoder is kept frozen and only the classifier operating on the image embedding $\mathbf{z}_i$ is trained. Two supervised classifiers are compared: LightGBM, as a machine-learning baseline, and a multilayer perceptron (MLP), as the neural classifier. LightGBM tests whether the ASEAN country classes are already separable by a non-neural model in the CLIP embedding space, while the MLP allows a learned nonlinear decision boundary.

The MLP configuration is selected by grid search on the validation set. The search varies the number of hidden layers, hidden width, dropout rate, activation function, and learning rate, using one or two hidden layers, 192 or 256 hidden units, dropout probabilities of 0.20 or 0.40, GELU or SiLU activation, and learning rates of $1 \times 10^{-3}$ or $5 \times 10^{-4}$. The selected MLP uses two hidden fully connected layers with 192 units per layer, SiLU activation, dropout probability 0.40, and a final linear layer that maps to 11 ASEAN country logits. Training uses cross-entropy loss with AdamW, weight decay of $1 \times 10^{-4}$, mini-batches of 128 samples, a maximum of 120 epochs, and early stopping with patience 12.

LightGBM is tuned on the same validation split over the number of estimators, number of leaves, and learning rate. The selected LightGBM uses 400 estimators, 31 leaves, a learning rate of 0.03, subsampling of 0.90, and column subsampling of 0.90. The final classifier configuration is selected by validation accuracy, with validation F1-score used as a secondary criterion. 

\subsection{Attention Map Generation}
Attention maps are generated after classification to provide a post-hoc visual explanation of the regions emphasized by the frozen CLIP vision encoder. This study uses attention rollout \cite{abnar_zuidema_2020} as the attention-based explanation method because it is directly compatible with Vision Transformer attention tensors and aggregates information across layers rather than relying on a single raw attention map. The resulting heatmaps are interpreted as spatial emphasis within the CLIP visual representation rather than direct class-specific attribution for the MLP output.

For each Transformer layer, attention weights are averaged across heads. An identity matrix is added to account for residual connections, and the resulting matrix is normalized row-wise:
\begin{equation}
\tilde{A}^{(l)}_{ij} =
\frac{A^{(l)}_{ij}+I_{ij}}
{\sum_k (A^{(l)}_{ik}+I_{ik})},
\end{equation}
where $\mathbf{A}^{(l)}$ denotes the head-averaged attention matrix at layer $l$, $\mathbf{I}$ is the identity matrix, and $\tilde{A}^{(l)}_{ij}$ is the row-normalized attention weight from token $i$ to token $j$. The final rollout matrix is obtained by recursively multiplying the normalized attention matrices across all $L$ layers:
\begin{equation}
\mathbf{R} =
\tilde{\mathbf{A}}^{(L)}
\tilde{\mathbf{A}}^{(L-1)}
\cdots
\tilde{\mathbf{A}}^{(1)}.
\end{equation}

For CLIP ViT, the rollout scores from the class token to the image patch tokens are reshaped into a two-dimensional patch grid and normalized to form an attention heatmap. The heatmap is resized to the image resolution for visualization and for comparison with object bounding boxes in the interpretability stage. Because CLIP preprocessing may resize or crop the original image before attention is computed, the spatial alignment between the resized heatmap and raw-image object boxes is interpreted as approximate.

\subsection{Object-Level Interpretability}
The attention maps described in the previous subsection provide spatial evidence at the patch or pixel level, but they do not directly indicate which semantic objects are associated with the prediction. To make the explanation more interpretable, this study compares the resized CLIP attention heatmap with object detections obtained from YOLO26. YOLO26 is applied to the original input image after the country prediction has been produced. The detector is not used by the classifier and does not receive the attention map as input; it only provides post-hoc bounding boxes, object class labels, and confidence scores. This design keeps the prediction pipeline separated from the interpretability pipeline: country prediction is based on frozen CLIP embeddings and the MLP classifier, whereas YOLO26 is used only to associate attended image regions with detected objects.

Let $\mathbf{M} \in \mathbb{R}^{H \times W}$ denote the resized attention map and let $b_k$ denote the bounding box of the $k$-th detected object. To evaluate object-level interpretability, we adapt energy-based pointing game metrics inspired by ScoreCAM \cite{wang2020scorecam}, where the score measures how much explanation energy falls inside a bounding box. This metric is also used for object-detector explanation evaluation in G-CAME \cite{nguyen2022gcame}:
\begin{equation}
\mathrm{EBPG}(b_k) =
\frac{\sum_{(x,y)\in b_k} \mathbf{M}(x,y)}
{\sum_{(x,y)} \mathbf{M}(x,y)}.
\end{equation}
This score represents the fraction of total attention energy that falls inside the detected object region. A higher EBPG value therefore indicates stronger overlap between the attention map and the object bounding box.

Raw EBPG can be biased toward large objects because larger bounding boxes occupy more image area and can accumulate more attention even when the attention density is not high. To reduce this area bias, this study additionally reports a complementary derived measure, called the attention density ratio (ADR):
\begin{equation}
\mathrm{ADR}(b_k) =
\frac{\mathrm{EBPG}(b_k)}
{|b_k|/(H W)},
\end{equation}
where $|b_k|$ is the bounding-box area. A value greater than one indicates that the object receives a higher attention density than expected from its image area alone, whereas a value below one indicates lower-than-expected attention density. We use EBPG as the primary energy-overlap metric and ADR as a secondary diagnostic for area bias. For each image, detected objects are ranked using this density ratio, and the scores can be aggregated by object class to identify recurring visual cues associated with correct predictions and failure cases. These values are interpreted as attention-object overlap measures rather than causal evidence, since they depend on the quality of YOLO26 detections, the granularity of bounding boxes, and the approximate spatial alignment between resized attention maps and raw input images.

\section{Result and Discussion}

\subsection{Country Classification}
Table~\ref{tab:country_classification_results} presents the country classification results on the held-out test set. The results indicate that task-specific supervised learning is important for this regional geolocation setting. CLIP zero-shot prediction provides a general vision-language baseline, but it is not adapted to the visual distribution of the ASEAN dataset. In contrast, LightGBM and MLP learn decision boundaries directly from the frozen CLIP embeddings using the training split, allowing them to emphasize dataset-specific visual patterns that separate countries with similar street-level scenes. This explains the improvement from 62.45\% accuracy for CLIP zero-shot to 82.82\% for LightGBM and 85.91\% for MLP. Since the dataset is approximately balanced, accuracy is used as the main comparison metric, while precision, recall, and F1-score are reported as supporting metrics. The MLP gives the best overall result and is therefore used as the final classifier for the subsequent interpretability analysis.

\begin{table}[!htbp]
    \centering
    \caption{Country classification performance on the held-out test set using frozen CLIP embeddings}
    \label{tab:country_classification_results}
    \footnotesize
    \setlength{\tabcolsep}{3pt}
    \begin{tabular}{lcccc}
        \toprule
        \textbf{Method} & \textbf{Accuracy} & \textbf{Precision} & \textbf{Recall} & \textbf{F1-score} \\
        \midrule
        CLIP zero-shot & 62.45\% & 65.69\% & 62.45\% & 61.27\% \\
        LightGBM & 82.82\% & 83.18\% & 82.82\% & 82.83\% \\
        MLP & \textbf{85.91\%} & \textbf{86.14\%} & \textbf{85.91\%} & \textbf{85.91\%} \\
        \bottomrule
    \end{tabular}
\end{table}

\begin{figure}[t]
    \centering
    \includegraphics[width=\columnwidth]{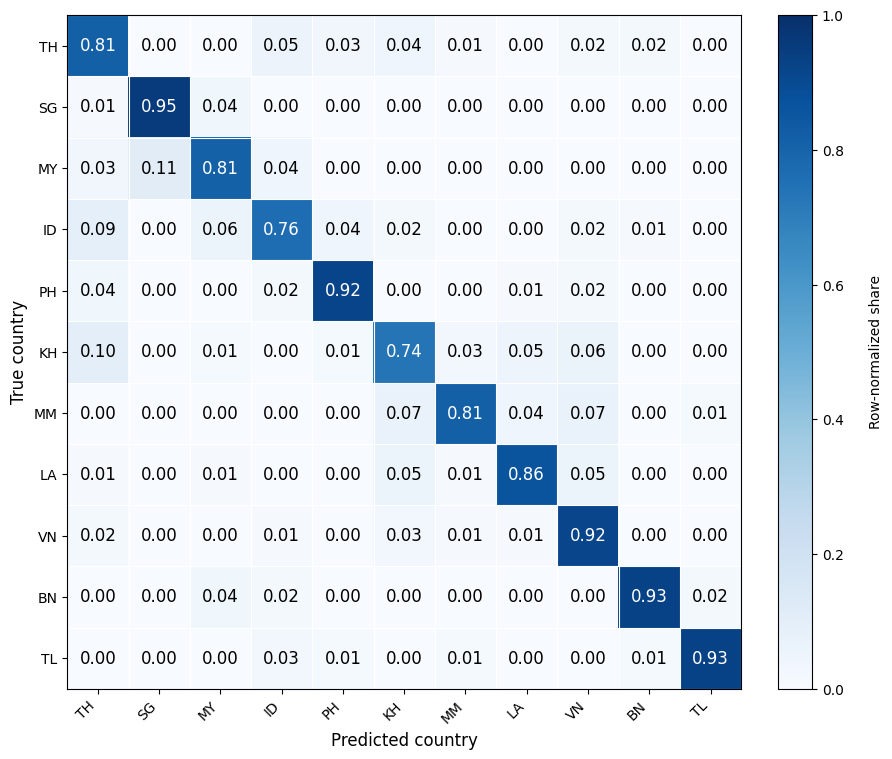}
    \caption{Confusion matrix of the country classifier on the held-out test set.}
    \label{fig:country_confusion_matrix}
\end{figure}

The confusion matrix in Fig.~\ref{fig:country_confusion_matrix} shows that Singapore, Brunei, Timor-Leste, the Philippines, and Vietnam are classified most reliably, with row-normalized diagonal values above 0.90. In contrast, Cambodia, Indonesia, Malaysia, Thailand, and Myanmar have lower diagonal values and more off-diagonal errors, suggesting that the classifier is less reliable when countries share similar roadside, commercial, architectural, or environmental cues.

Beyond the aggregate scores, the remaining errors are concentrated among visually and geographically related ASEAN countries rather than being uniformly distributed. The largest off-diagonal entries include Malaysia being predicted as Singapore (0.11), Cambodia as Thailand (0.10), and Indonesia as Thailand (0.09). The Malaysia--Singapore ambiguity is plausible because Singapore lies immediately south of Peninsular Malaysia, is directly connected to Johor, and shares dense commercial streetscapes with shophouses, covered walkways, shopfronts, and multilingual signage. The Cambodia--Thailand and Indonesia--Thailand errors suggest that the classifier sometimes treats roadside vegetation, markets, motorcycles, low-rise buildings, dense tropical urban streets, and regional signage textures as shared cues rather than country-specific evidence. Indonesia--Malaysia ambiguity is also consistent with shared tropical maritime street scenes and closely related Malay/Indonesian Latin-script signage when text is short, blurred, or partially visible. Overall, these patterns indicate that the MLP learns meaningful regional visual regularities, but still struggles when neighboring countries share environmental context, urban form, and partially visible text cues. The following subsections therefore inspect which cues are emphasized by the model.

\subsection{Attention Rollout Visualization}
\begin{figure}[t]
    \centering
    \includegraphics[width=\columnwidth]{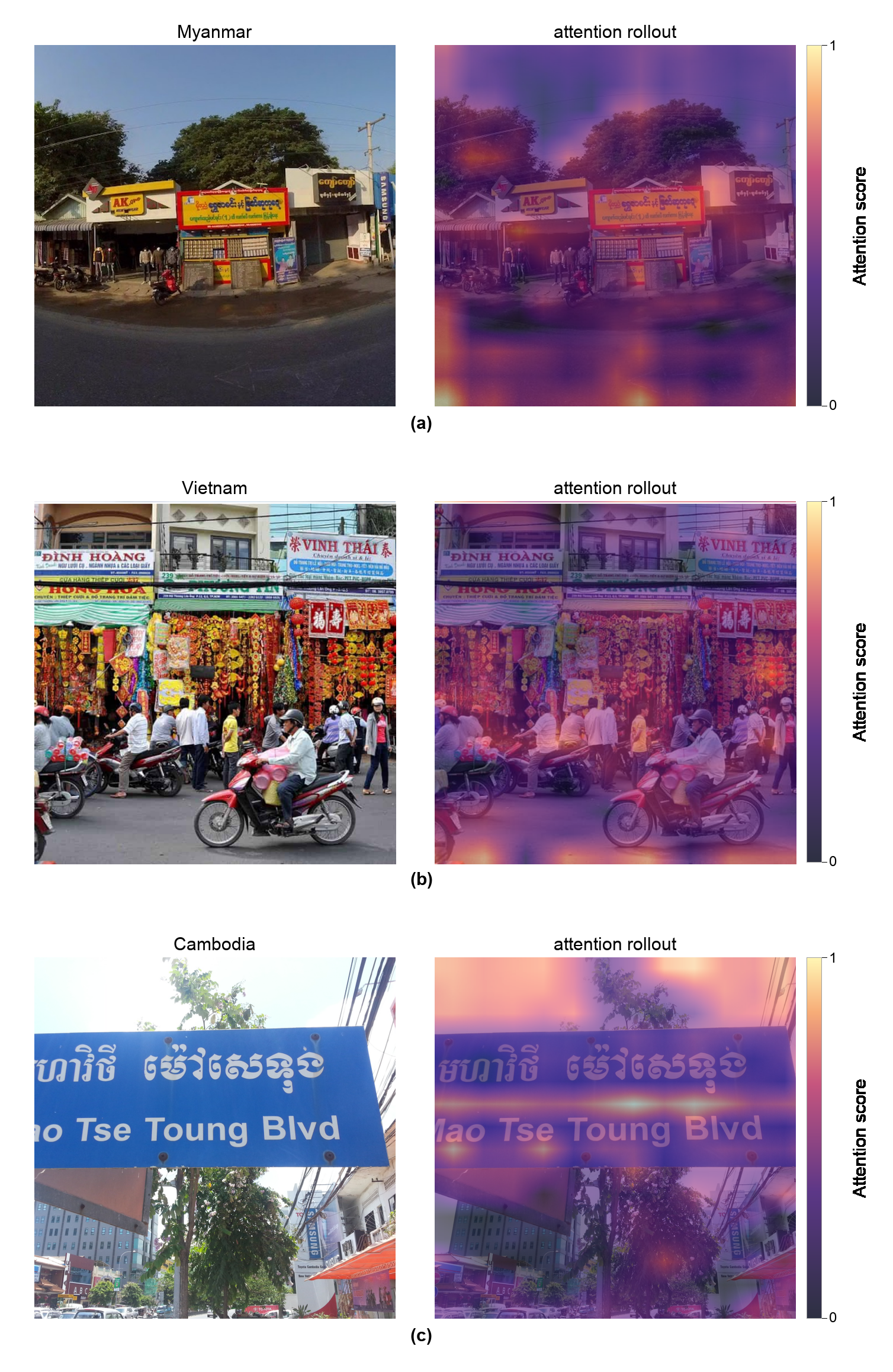}
    \caption{Examples of original test images and attention rollout heatmaps from the frozen CLIP vision encoder. Warmer regions indicate stronger spatial emphasis in the encoder representation and are later compared with YOLO26 detections for object-level analysis.}
    \label{fig:attention_rollout_examples}
\end{figure}

Attention rollout was used as a post-hoc visualization to inspect which image regions were emphasized by the CLIP vision encoder. In Fig.~\ref{fig:attention_rollout_examples}, the attention score is normalized from 0 to 1, where brighter yellow or orange regions indicate stronger attention and darker purple regions indicate weaker attention. To make the visual interpretation clearer, three country examples are inspected as representative cases:
\begin{itemize}
    \item \textbf{Myanmar:} Higher attention appears around the road foreground, storefront, and signage-bearing regions. This suggests that commercial facades, road structure, and distinctive non-Latin writing provide useful visual evidence for the Myanmar prediction.
    \item \textbf{Vietnam:} Attention is concentrated around dense shopfront signage, decorative storefront elements, motorcycles, pedestrians, and the road foreground. This pattern suggests that the model uses a busy commercial-street context, combining Vietnamese text cues with traffic and storefront structure rather than relying on a single object.
    \item \textbf{Cambodia:} The large road sign is partly emphasized together with surrounding sky, vegetation, wires, and roadside context. This indicates that visible script and environmental structure are interpreted together as part of the country-level evidence.
\end{itemize}
Overall, Fig.~\ref{fig:attention_rollout_examples} suggests that CLIP represents ASEAN street scenes through a mixture of text, storefronts, road layout, and environmental context. However, this visualization still requires manual inspection because the heatmap only shows emphasized regions and does not directly name the objects in those regions.

\subsection{Object Detection-Based Explanation}
\begin{figure}[t]
    \centering
    \includegraphics[width=\columnwidth]{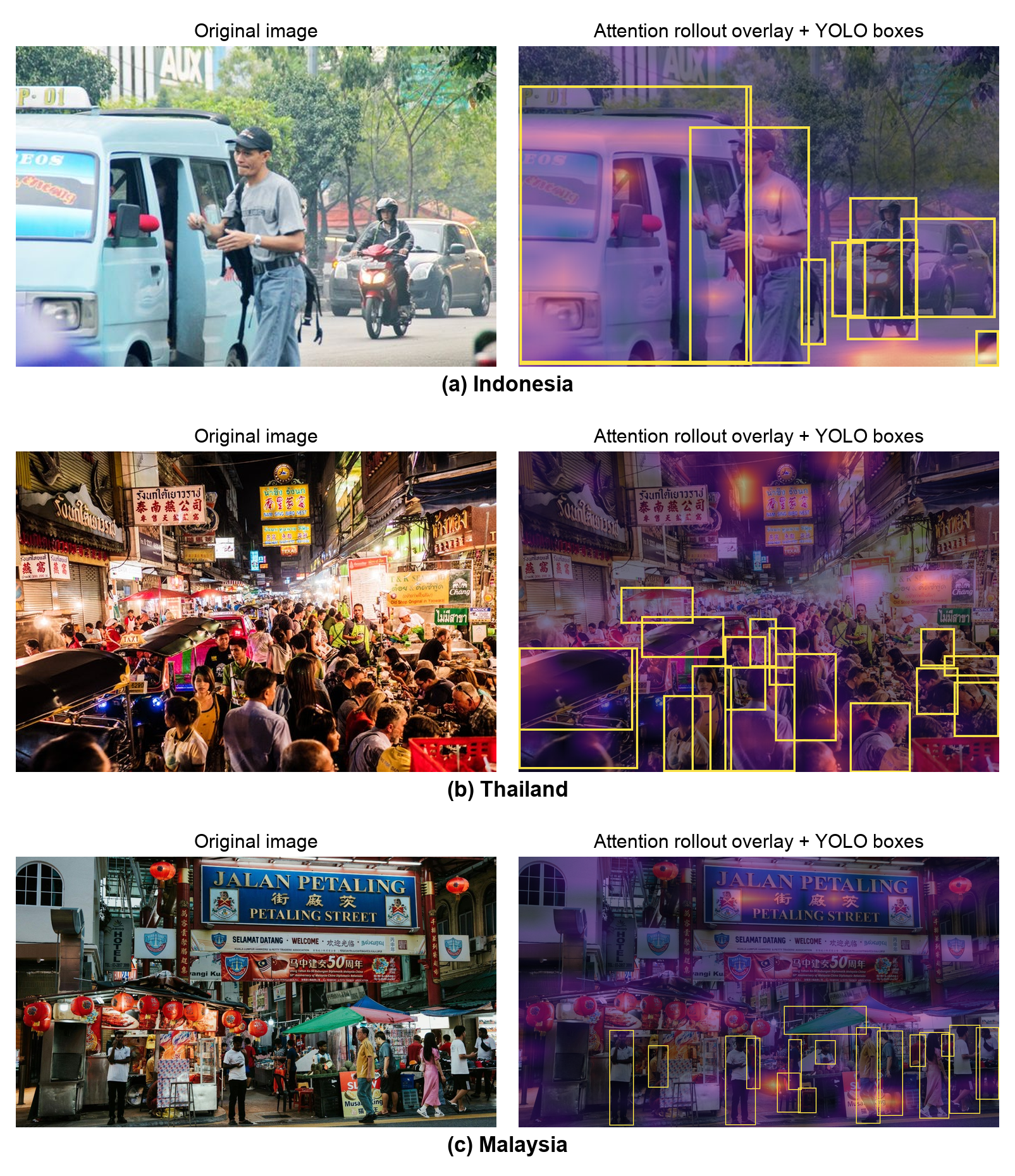}
    \caption{Examples of original test images and YOLO26 detections overlaid with CLIP attention rollout maps. The paired layout shows the raw visual context on the left and the detected objects associated with high-attention regions on the right.}
    \label{fig:ebpg_attention_overlays}
\end{figure}

To reduce the dependence on manual heatmap inspection, YOLO26 object detection was applied independently to the original images. The detector is not part of the classifier; instead, it provides post-hoc object labels and bounding boxes that can be compared with the CLIP attention rollout map. Fig.~\ref{fig:ebpg_attention_overlays} illustrates how detected objects overlap with high-attention regions, making it easier to describe which visible objects are associated with the model's spatial emphasis.

Several examples in Fig.~\ref{fig:ebpg_attention_overlays} show how object detection helps translate the attention map into more interpretable visual cues:
\begin{itemize}
    \item \textbf{Indonesia:} Attention appears around the vehicle body, nearby pedestrians, motorcycles, and roadside vegetation. This suggests that the image is represented through everyday traffic and roadside activity rather than a landmark-like object.
    \item \textbf{Thailand:} The attended regions are embedded in a crowded commercial street with dense signage, people, vehicles, night-market lighting, and shopfront structure. This makes the visual evidence closer to a street-level market texture than to a single detected object.
    \item \textbf{Malaysia:} Attention overlaps partly with pedestrians and umbrellas, but also spreads across Petaling Street signage, lanterns, shopfronts, and the surrounding commercial corridor. This suggests that object-level cues and broader commercial context are used together.
\end{itemize}
These cases show why object-level explanation is useful but incomplete. YOLO26 can name some attended objects, such as persons, vehicles, motorcycles, and umbrellas, while many geographically meaningful cues remain scene-level patterns such as signage density, storefront layout, lighting, road context, and commercial-street structure.

\subsection{Distribution of Detected Objects}
Beyond individual examples, the detected object distribution was analyzed to identify how object cues vary across countries. The goal is not to claim that one object class uniquely identifies one country, but to examine whether the same object categories appear with different prevalence and attention overlap in different ASEAN street contexts. In this sense, the object distribution provides a country-level comparison of traffic density, market activity, street furniture, commercial signage, and roadside infrastructure.

\begin{figure}[t]
    \centering
    \includegraphics[width=\columnwidth]{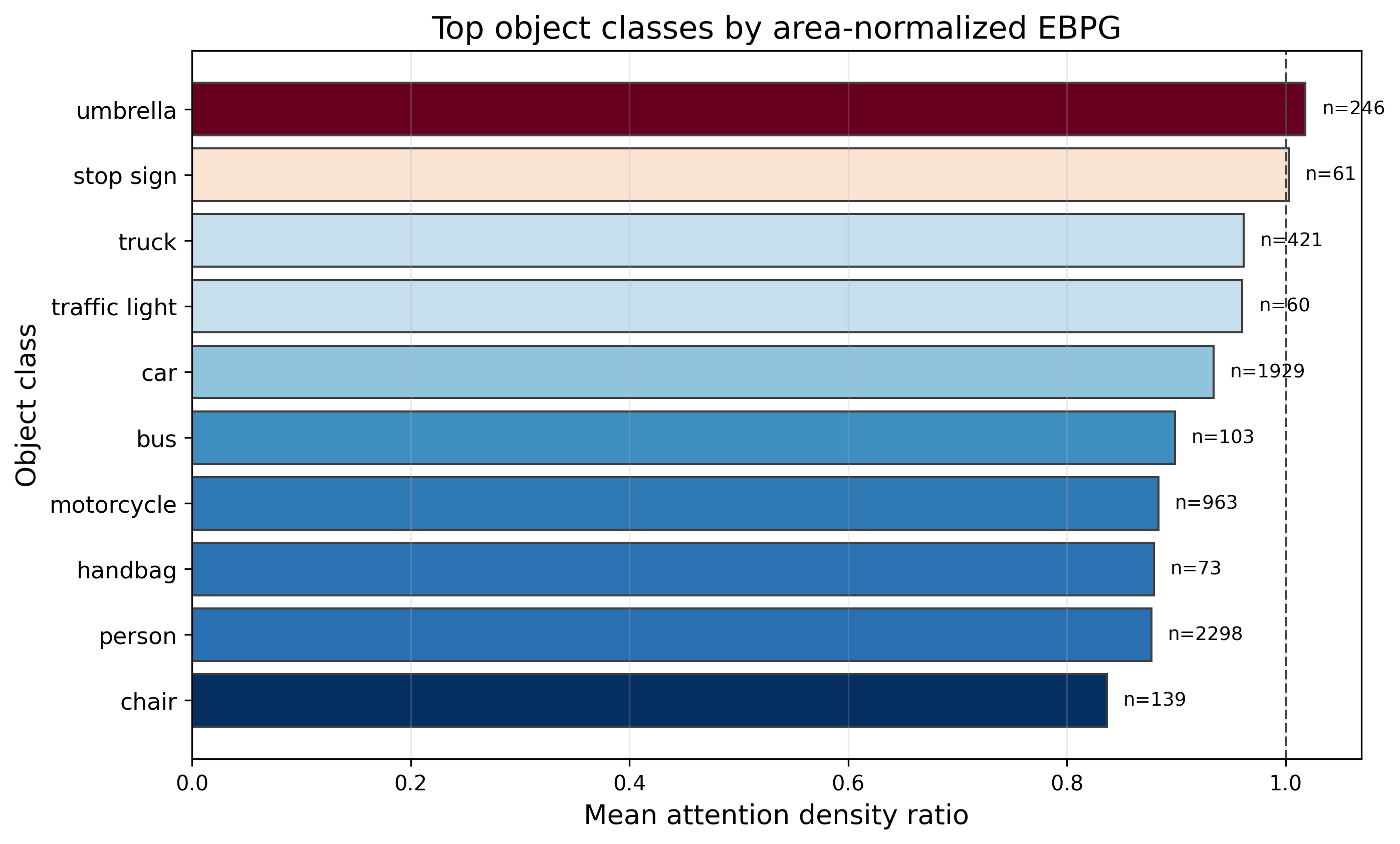}
    \caption{Top detected object categories ranked by area-normalized EBPG attention density ratio. Values above one indicate that the object receives higher attention density than expected from its bounding-box area.}
    \label{fig:ebpg_attention_density_objects}
\end{figure}

The aggregate results in Fig.~\ref{fig:ebpg_attention_density_objects} show that attention density is not driven only by object frequency. Persons, cars, and motorcycles are common in ASEAN street scenes, with 2,298, 1,929, and 963 detections, respectively, but their ADR values remain below one at 0.8772, 0.9343, and 0.8836. At the country level, however, their distributions are different. Motorcycles are most prominent in Vietnam, with 365 detections across 89 images, and are also frequent in Indonesia, Myanmar, and Cambodia. Cars are more dominant in Brunei, Laos, Malaysia, and Singapore, while motorcycles are rare in Singapore, Malaysia, and Brunei. This contrast suggests that the classifier may encounter different mobility patterns across ASEAN countries, even when the object labels themselves are generic.

Larger road objects also show country-dependent patterns. Truck detections are more frequent in the Philippines, Timor-Leste, Myanmar, Laos, and Indonesia than in Cambodia, Brunei, Malaysia, or Vietnam, while bus detections appear more often in the Philippines, Myanmar, Indonesia, and Timor-Leste. These differences should be interpreted as contextual evidence rather than fine-grained vehicle recognition: YOLO26 labels a bounding box as ``truck'' or ``bus'', but the geographic signal comes from how those vehicles appear together with road width, traffic density, roadside shops, signs, and surrounding built environments. Similarly, umbrellas have the highest ADR value at 1.0182 despite appearing only 246 times, and they are concentrated mainly in Cambodia and Laos in this subset, suggesting a link with shaded stalls, pedestrian activity, or market-like scenes. Stop signs remain close to the density baseline with 61 detections and appear across several countries, so they are better interpreted as traffic-infrastructure context than as a single-country marker. Overall, the distribution supports a compositional interpretation: ASEAN geolocation cues emerge from combinations of objects, signage, commercial activity, road infrastructure, and environmental context rather than isolated object categories.

\subsection{Failure Case Analysis}
\begin{figure}[t]
    \centering
    \includegraphics[width=\columnwidth]{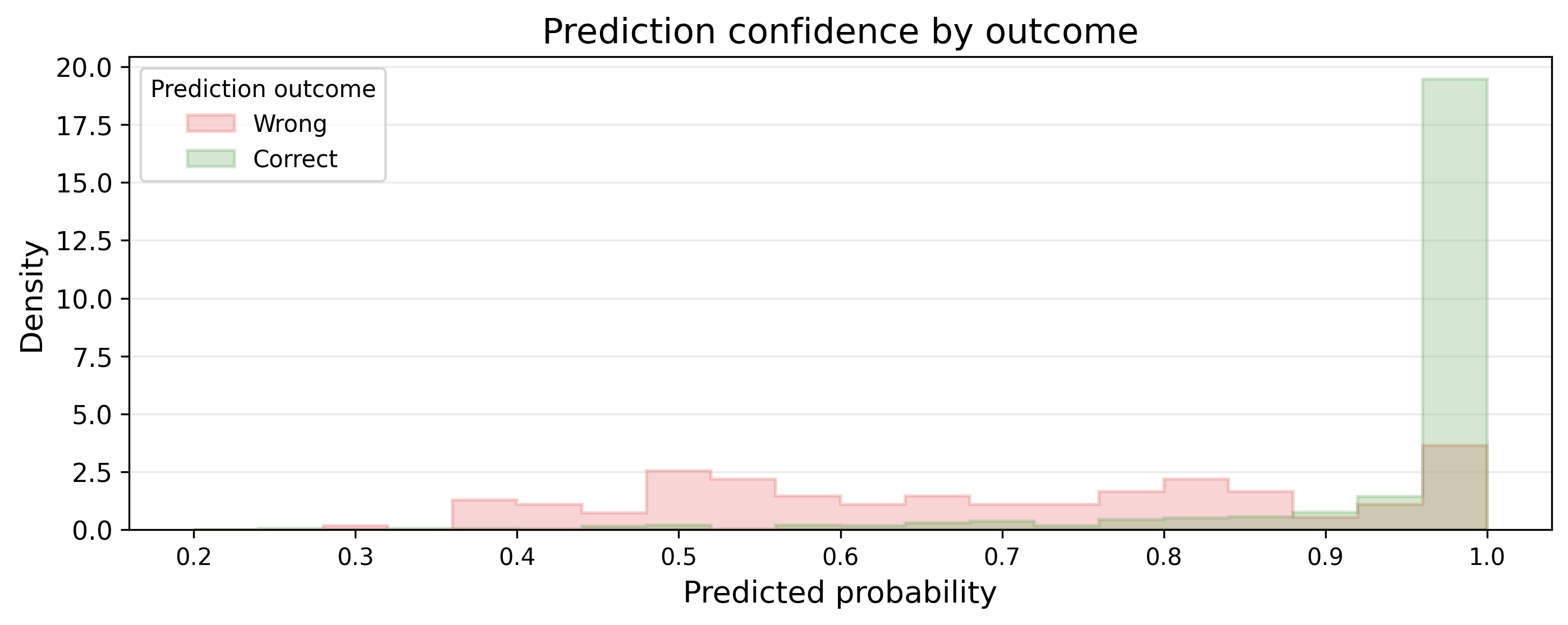}
    \caption{Prediction confidence for correct and incorrect test-set predictions. Incorrect predictions generally have lower confidence, but some failure cases remain highly confident.}
    \label{fig:confidence_correct_wrong}
\end{figure}

Failure cases were analyzed through the prediction-confidence distribution in Fig.~\ref{fig:confidence_correct_wrong}. The figure shows that correct predictions are strongly concentrated in the high-confidence region above 0.9, while incorrect predictions are more widely distributed across lower and mid-confidence values and still extend into the 0.8--0.9 range. This pattern suggests that a confidence score around 0.9 can be used as a practical high-confidence reference: predictions below this region are more uncertain and should be interpreted cautiously, especially when the classifier separates visually similar ASEAN countries.

However, confidence alone does not fully explain or resolve the failure cases. Some incorrect predictions also appear near or above the high-confidence region, indicating that the model can still be confident when it relies on visually plausible but misleading cues. These errors suggest that the main failure mode is not only low confidence, but visual ambiguity among countries with shared street-level patterns, including tropical roadside scenes, commercial areas, vegetation, vehicles, signage, and urban structures. Therefore, Fig.~\ref{fig:confidence_correct_wrong} supports using high confidence as a useful reliability cue, while also showing that visually ambiguous high-confidence errors remain an important limitation.

\section{Conclusion}
This study presented an explainable image geolocation pipeline for country-level prediction across 11 ASEAN countries by combining frozen CLIP image embeddings, a supervised classifier, and post-hoc visual explanation. The best classification performance was achieved by the MLP classifier, with 85.91\% accuracy on the held-out test set. Beyond accuracy, the explanation stage using CLIP attention rollout, YOLO26 object detection, and EBPG-based overlap analysis suggested that the model uses combinations of visual cues rather than isolated object categories. These results indicate that frozen vision-language representations can support accurate regional geolocation while still allowing object-level inspection of the visual evidence behind the prediction.

For future work, this study can be extended by comparing object detection-based explanations with image segmentation-based explanations to evaluate which approach provides better overlap with attention maps. Future studies should also expand the geographic scope beyond ASEAN to broader Asian regions and include more diverse urban, rural, and intra-country areas so that the dataset better represents visual variation within each country.

\end{document}